\documentclass[article,authoryear]{beg_32}             %
\usepackage[hang]{footmisc}
\usepackage[section]{placeins}   
\usepackage{longtable}          
\usepackage{hyperref} 
\makeatletter
\@ifclasswith{beg_32}{review}{%
  \let\beg@maketitle\maketitle
  \renewcommand{\maketitle}{%
    \edef\beg@savedstretch{\baselinestretch}%
    \setstretch{1}\beg@maketitle\setstretch{\beg@savedstretch}\normalsize}%
  \AtBeginDocument{%
    \renewcommand{\normalsize}{\fontsize{12}{14.4}\selectfont}%
    \renewcommand{\small}{\fontsize{11}{13.2}\selectfont}%
    \captionsetup[figure]{font=normalsize}%
    \captionsetup[table]{font=normalsize}%
    \setlength{\emergencystretch}{3em}%
    \normalsize}%
  \usepackage{etoolbox}%
  \AtBeginEnvironment{tabular}{\fontsize{10}{12}\selectfont}%
  \AtBeginEnvironment{longtable}{\fontsize{10}{12}\selectfont}}{}
\makeatother
\fancypagestyle{plain}{%
  \fancyhf{}
  \fancyfoot[R]{\small\bf\thepage }
  }
\renewcommand{\dmyy}{20}
\renewcommand{\myyear}{yyyy}
\renewcommand{\today}{}
\makeatletter
\@ifclassloaded{beg_32}{}{%
  \usepackage{authblk}%
  \def\address{\affil}%
  \def\volume#1{}%
  \providecommand{\titlehead}[1]{}%
  \providecommand{\authorhead}[1]{}%
  \providecommand{\dataO}[1]{}%
  \providecommand{\dataF}[1]{}%
  \providecommand{\corremail}[1]{}%
  \providecommand{\corraddress}[1]{}%
  \providecommand{\corrauthor}[2][]{\author[#1]{#2}}%
  \providecommand{\titlecap}[1]{#1}}
\makeatother
\begin{document}

\volume{Volume x, Issue x, \myyear\today}
\title{Fractional physics-informed neural networks to model social media virality}
\titlehead{Fractional PINNs for Social Media Virality}
\authorhead{A. Koong \& G. E. Karniadakis}
\author[1]{Alex Koong}
\corrauthor[2]{George Em Karniadakis}
\corremail{george\_karniadakis@brown.edu}
\corraddress{Brown University, Division of Applied Mathematics, Providence, RI, 02906, USA}
\address[1]{Brown University, Division of Applied Mathematics, Providence, RI, 02906, USA; E-mail: alexander\_koong@brown.edu}
\address[2]{Brown University, Division of Applied Mathematics, Providence, RI, 02906, USA; E-mail: george\_karniadakis@brown.edu}

\dataO{}
\dataF{}

\abstract{Existing models of social media virality either prescribe the functional form of engagement growth or assume a fixed mechanism by which past activity influences future engagement. We present a fractional physics-informed neural network that learns both the temporal memory and nonlinear growth dynamics of social media engagement, using symbolic regression to translate the learned dynamics into interpretable equations. Synthetic experiments at the deployment resolution of 24 hourly observations demonstrate that the fractional order and reaction law can be jointly recovered from clean data. Under observation noise, however, the standard pointwise physics residual drives the fractional order toward its lower bound through a scale artifact. We address this failure using a fully whitened residual that propagates observation noise through the fractional operator. Applying the framework to 348 posts from X (formerly Twitter), we find that a single-window model estimates strongly memory-dependent dynamics but fails to accurately reconstruct held-out trajectories. Partitioning the 24-hour window into four teacher-forced segments substantially improves performance, yielding segment-specific fractional orders that increase from 0.23 to 0.99 and interpretable growth laws for each phase. The segmented model reconstructs 64\% of held-out trajectories within a mean squared error of 0.01, compared with 44\% for the strongest of nine classical baselines. Together, these results demonstrate the potential of fractional physics-informed equation discovery to uncover interpretable engagement dynamics while highlighting the importance of noise-aware estimation and temporal segmentation.}


\maketitle

\section{Introduction}
Physics-informed neural networks (PINNs), introduced by \citet{raissi2019pinn}, embed known physical laws directly into the training of a neural network. Rather than learning an input-output map from data alone, PINNs minimize loss functions that combine a data-fitting term with a physics residual term. The data-fitting term measures the error between predicted and observed values, while the physics residual term measures how far a prediction deviates from a governing differential equation. The two work in tandem and regularize each other; the data term anchors the solution to observations, while the physics residual keeps the network consistent with the governing equation. PINNs generally outperform standard neural networks in regimes with sparse or noisy data, hence their popularity in modeling physical systems such as fluid dynamics, materials science, and geophysics \citep{karniadakis2021physics}.

A more recent line of work uses PINNs not to solve known equations (the \textit{forward problem}), but to discover unknown equations and parameters (the \textit{inverse problem}). When the governing law is not specified in advance, the unknown term can be represented by a neural network during training and later distilled into a closed-form expression. This distillation is usually done with symbolic regression tools such as PySR \citep{cranmer2023pysr}. \citet{zhang2024discovering} applied this idea to computational neuroscience and recovered the reaction term of a network model for the spread of misfolded tau proteins across the brain in Alzheimer's disease. Their model takes the form
\begin{equation}
\frac{dc}{dt} = -\kappa L c + \alpha f(c),
\end{equation}
where $c$ is the vector of tau concentrations across brain regions, $L$ is the graph Laplacian of the structural connectome, $\kappa$ is the diffusive coupling, and $f(c)$ is the unknown reaction function. By solving the inverse problem, they recovered a polynomial expression for $f(c)$ and thereby a governing law for the spread of misfolded tau proteins.

Although PINNs have been used to solve inverse problems across a wide range of physical science domains, they have yet to be fully explored in the social sciences. Many social processes share two key features with physical ones: they are governed by some universal mechanism, and the data that describe them are often sparse or noisy. Social media virality is one example. The spread of content is shaped by both human behavior and a platform-specific content algorithm, while observations of how any given post spreads are typically limited. This makes it a natural candidate for PINNs. In this study, we present a novel application of PINNs: discovering an unknown governing equation for social media virality on X (formerly Twitter). To the best of our knowledge, this is the first use of PINNs to solve an inverse problem in this domain.

\section{Problem Definition}
\label{sec:ProblemDef}
Social media virality, defined as the spread of online content, is generally thought of as a cascade process: a viewer engages with content, which in turn surfaces the content to other viewers, who additionally engage with the content and complete the loop. This sequential chain reaction continues until new viewers stop engaging, which breaks the loop and virality levels out. Virality is typically measured in units of ``engagements.'' Depending on the social media platform, this is a combination of views, likes, comments, reposts, or any other platform-specific metrics. For the remainder of this paper, virality on X is measured by total engagements, which we define to be the sum of engagements a post receives (likes, comments, reposts, quote posts). Each form of engagement is weighted equally.

Existing approaches to modeling virality generally fall into one of two families: growth-curve models and self-exciting point processes. Growth-curve models, including the logistic \citep{verhulst1838notice}, Gompertz \citep{gompertz1825nature}, and Bass diffusion \citep{bass1969diffusion}, describe cumulative engagement with a small set of interpretable parameters, but commit to a functional form before seeing any data. The logistic model, for example, fixes growth to be proportional to $c(1-c)$, so its highest growth rate must occur exactly when a post reaches half its eventual size. Posts that deviate from this shape cannot be fit well, no matter how the parameters are tuned. Additionally, growth-curve models are first-order and Markovian, meaning for a fixed set of parameters, the future trajectory depends only on current engagement, not on how that engagement was reached. Two posts with the same current engagement but different early-hour growth would therefore receive identical forecasts, leaving valuable information in the post's history unused.

In contrast, self-exciting point processes such as Hawkes processes \citep{hawkes1971spectral} and the SEISMIC model \citep{zhao2015seismic} treat every interaction as raising the intensity of future interactions, so the full history of a post matters. However, the memory kernel that governs how quickly this influence decays must still be specified in advance, making these models poorly suited to regimes where the governing law is not known beforehand. Fitting them also requires event-level timestamps, which are often unavailable in practice. Taken together, these two families of models hard-code the two ingredients that should be learned, not assumed, when modeling a process as complex as social media virality. This gap in the ability to learn ``what form does viral growth take'' and ``how strongly does a post's history drive its future'' directly motivates our research.

We propose using PINNs to bridge this modeling gap. In our method, virality is represented as a differential equation with two unknown components that are recovered during training. The growth law is represented by a neural network and is later distilled into a closed form. The weight of history is represented by a fractional derivative whose order $\beta$ describes the dynamics of memory, following the fractional PINN (fPINN) framework of \citet{pang2019fpinns}, in which the fractional operator is discretized inside the physics residual. Our contributions are as follows. First, we present a fractional PINN framework for social media engagement that discovers its own reaction dynamics. Second, we use synthetic data to validate the model's ability to recover unknown parameters. Third, we train the model on real-world data. Finally, we compare our method with classical baselines and discuss future directions for applying PINNs to other areas of social science.

\section{Methods}
\label{sec:Methods}

\subsection{Model Design}
\subsubsection{Governing equation}
We define the following fundamental ``physical'' law of social media: the rate of change of a post's engagement, computed from its full history, is equal to the post's viral potential multiplied by some universal growth law. This is represented by the following equation:
\begin{equation}
  D_t^{\beta} c_i(t) = \alpha_i\, f\!\left(c_i(t)\right), \qquad c_i(0)=c_{i,0},
  \label{eq:model}
  \end{equation}
where $D_t^{\beta} c_i(t)$ is the rate of change of post $i$'s engagement, $\alpha_i$ is the post's viral potential, and $f(c_i(t))$ is a universal growth law. $\alpha_i$ is unique to each individual post, whereas $f$ is learned from data and shared across all posts. $c_i(t)\in[0,1]$ represents the normalized cumulative engagement of post $i$ and is modeled as the solution of the fractional initial value problem.

\subsubsection{Caputo fractional derivative}
\label{sec:caputo}

The operator $D_t^{\beta}$ in Eq.~\eqref{eq:model} is the Caputo fractional derivative \citep{caputo1967linear} of order $\beta\in(0,1)$. On the rescaled interval $t\in[0,1]$, it is defined as
\begin{equation}
D_t^{\beta} c(t) = \frac{1}{\Gamma(1-\beta)} \int_0^t \frac{c'(\tau)}{(t-\tau)^{\beta}}\, d\tau .
\label{eq:caputo}
\end{equation}
Unlike an integer-order derivative, which depends only on the instantaneous state, the Caputo derivative integrates over the entire prior trajectory, weighting recent changes more heavily than distant ones through the power-law kernel $(t-\tau)^{-\beta}$. $\beta$ sets the overall rate of decay and represents how quickly a post's past engagement is forgotten: as $\beta \rightarrow 0$, a post's history has a stronger influence in shaping future virality. Because $\beta$ is learned from data and shared across all posts, the model infers the memory strength of social media engagement rather than fixing it in advance. During training, $\beta$ is parameterized as $\beta = 0.1 + 0.89\,\sigma(\theta_\beta)$ with $\theta_\beta$ unconstrained, which keeps it inside $(0.1, 0.99)$. We evaluate the derivative on the discrete hourly observations using the L1 scheme \citep{lin2007finite} on a uniform grid $t_n = nh$, $n = 0, \dots, T-1$:
\begin{equation}
D_t^{\beta} c(t_n) \approx \frac{h^{-\beta}}{\Gamma(2-\beta)} \sum_{j=0}^{n-1} b_j^{(\beta)}\, \big(c_{n-j} - c_{n-j-1}\big),
\label{eq:l1}
\end{equation}
with weights $b_j^{(\beta)} = (j+1)^{1-\beta} - j^{1-\beta}$.

\subsubsection{Per-post scaling factor}

The per-post scaling factor, or \emph{viral potential} $\alpha_i$, sets the amplitude of the growth dynamics for each post. Each post has a unique $\alpha_i$ that is computed from two observable covariates: the author's follower count $N_i$ and the post's early engagement velocity $v_i$. Viral potential $\alpha_i$ is defined as
\begin{equation}
\alpha_i = \sigma\!\big(a\,\log(1+N_i) + b\,v_i + d\big), \qquad v_i = \log\!\big(1+E_i(1)\big) - \log\!\big(1+E_i(0)\big),
\label{eq:alpha}
\end{equation}
where $\sigma$ is the logistic sigmoid, $E_i(0)$ and $E_i(1)$ are post $i$'s raw cumulative engagement at its first two observed hours, and $(a, b, d)$ are scalars shared across all posts and learned during training. We combine the two covariates because follower count alone carries too little signal to predict how quickly a post reaches its peak engagement. The sigmoid keeps $\alpha_i$ in $(0,1)$ while remaining smooth and differentiable everywhere.

\subsubsection{Reaction network and training}
\label{sec:training}

The reaction law is represented by a four-layer fully connected neural network of width 30 whose output is $f(c) = e^{\mathrm{NN}(c)}\, c\,(1-c)$, which enforces the boundary conditions $f(0) = f(1) = 0$ and keeps $f$ strictly positive on $(0,1)$. All unknowns (the network weights, $\theta_\beta$, and the viral-potential coefficients $(a,b,d)$) are trained jointly with the Adam optimizer \citep{kingma2015adam}. The model is implemented in JAX and trained on a CPU.

\subsection{Synthetic Data Validation}\label{sec:synthetic}

Before applying the model to real engagement data, we validate that the inverse problem is actually solvable: given trajectories generated by the governing equation with known parameters, does the training procedure recover those parameters? We proceed one unknown at a time: first verifying the numerical schemes themselves, then recovering the fractional order $\beta$ in isolation, then recovering $\beta$ and the reaction law $f(c)$ jointly, and finally testing the robustness of the joint recovery on perturbed synthetic data. The per-post potential is held at its ground-truth value ($\alpha = 1$) throughout this section, as neither $\beta$'s identifiability nor $f$'s shape depends on what $\alpha$ is fixed to.

\subsubsection{Synthetic data generation}

All experiments share the same clean synthetic data, generated from the ground-truth equation
\begin{equation}
  D_t^{0.5} c(t) = 1.0 \cdot c(1 - c^2), \qquad c(0)=c_0.
  \label{eq:ground_truth}
\end{equation}
We use this cubic reaction law with $f(0)=f(1)=0$ to generate 200 trajectories with $c_0 \sim \mathcal{U}(0.01, 0.3)$ on the rescaled interval $t \in [0,1]$. Each trajectory is observed at $T = 24$ uniform nodes, matching the 24 hourly observations per post in the real data. For the $\beta$-recovery experiment we repeat everything at ground-truth $\beta \in \{0.3, 0.7\}$, so that claims of accuracy are not tied to a single point in parameter space.

Each trajectory is solved with the Adams--Bashforth--Moulton predictor--corrector for fractional differential equations \citep{diethelm2002predictor} on a fine grid of approximately 8,192 points and then subsampled to the $T$ observation nodes. The solver converges at its theoretical rate of $O(h^{1+\beta})$ against the exact Mittag-Leffler solution of the linear problem $D_t^{\beta} c = c$ (observed orders $1.30$, $1.49$, and $1.78$ for $\beta = 0.3, 0.5, 0.8$). Tripling the fine grid to approximately 24,576 points changes the observed values by at most $3.8\times10^{-6}$ ($\beta=0.3$; $2.8\times10^{-7}$ and $2.6\times10^{-8}$ for $\beta = 0.5, 0.7$). The synthetic observations are therefore numerically exact for all practical purposes.

\subsubsection{Verification of the L1 discretization of the Caputo derivative}

The training loss evaluates the Caputo derivative of observed data with the L1
scheme of Eq.~\eqref{eq:l1}, so we first verify this scheme against test
functions with known Caputo derivatives: $c(t) = t^{2}$ and $c(t) = \sin(\pi t)$. Across grid resolutions $T \in \{24, 48, 96, 192, 384\}$ the scheme converges at its theoretical rate of $O(h^{2-\beta})$ \citep{lin2007finite}. Observed orders are $1.68$, $1.50$, and $1.20$ for $\beta = 0.3, 0.5, 0.8$, against theoretical values $1.7$, $1.5$, $1.2$.  Maximum errors at the working resolution $T=24$ with $\beta=0.5$ are $4.1\times10^{-3}$ for $t^{2}$ and $1.9\times10^{-2}$ for $\sin(\pi t)$. The L1 scheme is thus correctly implemented and converges as theory predicts.

\subsubsection{Recovery of the fractional order $\beta$ alone}

We first ask whether $\beta$ can be recovered when everything else is known.
With $f$ and $\alpha$ fixed at their true values, the physics loss reduces to
a scalar function of $\beta$. We scan the loss on a dense grid over $\beta \in [0.11, 0.95]$ and refine its minimizer by bounded scalar optimization.  In every setting the landscape is smooth with a single well-separated minimum, so the results below are properties of the objective itself rather than of any particular optimizer.

Because the L1 scheme is least accurate at the earliest time nodes for
solutions of fractional initial value problems, we evaluate two
formulations of the physics residual: the pointwise L1 residual of
Eq.~\eqref{eq:l1} over all interior nodes, and the same residual excluding
the first four interior nodes (burn-in).
Table~\ref{tab:beta_recovery} reports the recovered $\hat{\beta}$ at the
working resolution $T = 24$.  Both formulations recover $\beta$ to within
$0.028$ for all three ground truths.  The pointwise L1 residual carries a
small systematic downward bias of approximately $-0.025$; the burn-in
variant reduces this bias by a factor of at least $5$, to at most $0.005$,
confirming that most of the pointwise error originates at the earliest
nodes. The fractional order is therefore identifiable, despite irreducible discretization error.

\begin{table}[!htb]
\centering\begin{threeparttable}

\caption{Recovered $\hat{\beta}$ at $T = 24$ for each residual formulation
(signed error in parentheses).}
\label{tab:beta_recovery}
\centering\begin{tabular}{|ccc|}
\hline
\textbf{True $\boldsymbol{\beta}$} & \textbf{Pointwise L1} & \textbf{L1, burn-in} \\
\hline
0.3 & 0.2751 ($-0.0249$) & 0.2954 ($-0.0046$) \\
0.5 & 0.4729 ($-0.0271$) & 0.4965 ($-0.0035$) \\
0.7 & 0.6779 ($-0.0221$) & 0.7002 ($+0.0002$) \\
\hline
\end{tabular}
\end{threeparttable}
\end{table}

\subsubsection{Joint recovery of $\beta$ and $f(\mathit{c})$}

Next, we free the reaction law and ask whether the model can jointly recover $\beta$ and $f(c)$. The reaction law is represented by the network of Section~\ref{sec:training}, and $\beta$ is learned through the sigmoid parameterization of Section~\ref{sec:caputo}. For each of the two residual formulations (pointwise L1, burn-in L1) we train four independent runs (two $\beta$ initializations and two network seeds), each for 30,000 Adam steps on the same 200-trajectory dataset.

Table~\ref{tab:joint_recovery} shows that joint recovery is stable.  Under the pointwise L1 residual, all four runs
converge to numerically indistinguishable solutions regardless of
initialization or seed. The spread in $\hat{\beta}$ across runs is
$2.3\times10^{-4}$ ($4.2\times10^{-4}$ for the burn-in variant), two orders
of magnitude below the biases being measured.  Averaged over runs, the
pointwise residual recovers $\hat{\beta} = 0.457$, a correlation between the
learned and true $f$ of $0.9978$ on $[0,1]$, and a learned $f$ amplitude
within $0.7\%$ of the truth.  Relative to the $\beta$-only experiment, the
downward bias of the pointwise residual widens from $-0.027$ to $-0.043$. We attribute this to the flexible $f$ absorbing part of the early-node residual error, while the shape and scale of $f$ remain essentially exact.  The burn-in variant again
removes most of the bias: its runs converge to $\hat{\beta} = 0.494$ with an
$f$ correlation of $1.0000$ and an amplitude within $0.4\%$ of the truth.

\begin{table}[!htb]
\centering\begin{threeparttable}

\caption{Joint recovery of $\beta$ and $f(c)$ on clean data at $T = 24$
(true $\beta = 0.5$; mean over the four runs of each formulation; signed
$\hat{\beta}$ error in parentheses).}
\label{tab:joint_recovery}
\centering\begin{tabular}{|cccc|}
\hline
\textbf{Residual} & \textbf{Recovered $\boldsymbol{\hat{\beta}}$} & \textbf{$\boldsymbol{f}$ correlation} & \textbf{$\boldsymbol{f}$ amplitude error} \\
\hline
Pointwise L1 & 0.4574 ($-0.0426$) & 0.9978 & $+0.7\%$ \\
L1, burn-in  & 0.4945 ($-0.0055$) & 1.0000 & $+0.4\%$ \\
\hline
\end{tabular}
\end{threeparttable}
\end{table}

We then distill the learned $f$ from the best run of each formulation into a
closed-form expression with PySR \citep{cranmer2023pysr}, restricted to the
operators $\{+, -, \times, \div, \exp\}$ at maximum complexity 14. We recover
\begin{equation}
\hat{f}(c) = c\,\big(0.9914 + 0.0403\,c - 1.0320\,c^{2}\big),
\label{eq:pysr_recovered}
\end{equation}
a near-identical cubic to the ground-truth law (correlation $0.99998$
with $c(1-c^{2})$).

Taken together, these experiments show that the model recovers the full
unknown structure of Eq.~\eqref{eq:model} --- the fractional order, the
reaction law, and its closed form --- from trajectories at the deployment
resolution.  All of the above, however, rests on numerically exact
observations.  In the next section we perturb the synthetic observations and show that noise plays a large role in whether recovery is successful.

\subsubsection{Joint recovery of $\beta$ and $f(\mathit{c})$ with perturbed data}
\label{sec:perturbed}

The last step of the model verification is to ensure that the model can recover the correct parameters when trained on perturbed synthetic data. We perturb the same 200 trajectories with observation noise such that $c^{\mathrm{obs}} = \mathrm{clip}(c + \varepsilon, 0, 1)$. For each node, $\varepsilon \sim \mathcal{N}(0, \sigma^{2})$ is drawn independently, with $\sigma$ ranging from $0.005$ to $0.05$. Two independent noise realizations are generated for each $\sigma$. All other steps remain the same.

Unlike with the clean synthetic data, the pointwise objective fails to recover $\beta$ at even the smallest perturbation: $\hat{\beta}$ slides downward and pins at the lower bound of its parameterization ($\beta = 0.1$) by $\sigma = 0.02$. The cause is a structural flaw in the loss itself, illustrated in Figure~\ref{fig:loss_cheat}: the residual carries a $\beta$-controlled scaling term, so the optimizer shrinks the objective by shrinking the \emph{scale} of the error rather than the error itself. At $\sigma = 0.02$ the trained model attains an objective value seven times smaller than the true parameters attain on the same data.  Excluding early nodes (burn-in) does not help because the noise floor lives at every node.

\begin{figure}[!htb]
\centering
\begin{displaymath}
\mathcal{L}_{\mathrm{pointwise}} \;=\; \frac{1}{m}\sum_{n=1}^{m}
\Bigg(\; \underbrace{\frac{h^{-\beta}}{\Gamma(2-\beta)}}_{\text{operator gain } g(\beta)}
\,\sum_{j=0}^{n-1} b_j^{(\beta)}\big(c^{\mathrm{obs}}_{n-j}-c^{\mathrm{obs}}_{n-j-1}\big)
\;-\;\underbrace{\alpha\, f\!\big(c^{\mathrm{obs}}_{n}\big)}_{\text{reaction}} \Bigg)^{\!2}
\end{displaymath}
\caption{The pointwise objective. The braced factor is the operator gain
$g(\beta) = h^{-\beta}/\Gamma(2-\beta)$. Because $D^{\beta}c$ carries units of
$c\,t^{-\beta}$, $g$ is the unit in which the residual is measured: at
$h = 1/23$, $g$ falls from ${\approx}\,22$ at $\beta = 0.99$ to
${\approx}\,1.4$ at $\beta = 0.1$. Once observation noise dominates the
increments, the residual is proportional to $g(\beta)$, so lowering $\beta$ alone shrinks the loss with no improvement in fit.}
\label{fig:loss_cheat}
\end{figure}

We first attempt to fix this with \emph{diagonal whitening}. The residual itself
is unchanged and uses the same L1 discretization and reaction term; we simply change the units in which each residual is measured. Under the noise model, the residual $r_n$ is not zero even at the true
parameters. Propagating the observation noise through the residual gives
it a predicted standard deviation of $\hat{\sigma}\sqrt{v_n}$, where $v_n$
collects the squared sensitivities of $r_n$ to each observation it
touches. Dividing each squared residual by this predicted variance turns
every term of the loss into a squared $z$-score, which closes the exploit described above: the gain $g(\beta)$ multiplies the data increments in $r_n$ and cancels in the ratio. In other words, lowering $\beta$ no longer lowers the loss, and the optimizer can only decrease the objective by making the residuals small relative to the noise. This is the errors-in-variables view of the regression problem \citep{fuller1987measurement}: the observations, not the model, carry the noise. The full objective is
\begin{equation}
\mathcal{L}_{\mathrm{diag}} \;=\; \frac{1}{m}\sum_{n=1}^{m}\left[\frac{r_n^{2}}{\hat{\sigma}^{2}\, v_n} \;+\; \log v_n\right],
\qquad v_n \;=\; \sum_{k} J_{nk}^{2},
\label{eq:diag_whitened}
\end{equation}
where $r_n$ is the same pointwise residual as before:
\begin{equation}
r_n \;=\; \underbrace{\frac{h^{-\beta}}{\Gamma(2-\beta)}
\sum_{j=0}^{n-1} b_j^{(\beta)}\big(c^{\mathrm{obs}}_{n-j}-c^{\mathrm{obs}}_{n-j-1}\big)}_{\text{L1 derivative at node } n}
\;-\;
\underbrace{\alpha\, f\!\big(c^{\mathrm{obs}}_{n}\big)}_{\text{reaction}} .
\label{eq:diagwhitenpart2}
\end{equation}
Here $\hat{\sigma}$ is a robust noise estimate taken from second differences of the data, and the $\log v_n$ term keeps the optimizer from simply inflating the predicted variances. $v_n$, the variance the noise model predicts for the $n$-th residual, comes from the Jacobian of the residual vector $r = (r_1, \dots, r_m)$ with respect to the observations,
\begin{equation}
J \;=\; \frac{\partial r}{\partial c^{\mathrm{obs}}} \;=\; A(\beta) \;-\; \alpha\,\mathrm{diag}\!\big(f'(c^{\mathrm{obs}})\big),
\label{eq:jacobian}
\end{equation}
where $\mathrm{diag}\!\big(f'(c^{\mathrm{obs}})\big)$ denotes the $m\times(m+1)$
matrix with $f'\!\big(c^{\mathrm{obs}}_{n}\big)$ in entry $(n,n)$ for
$n = 1,\dots,m$ and zeros elsewhere (in particular, a zero column at $k = 0$),
and $A(\beta)$ is the matrix form of the L1 operator.

\begin{figure}[!htb]
	\centering
  \includegraphics[width=13cm]{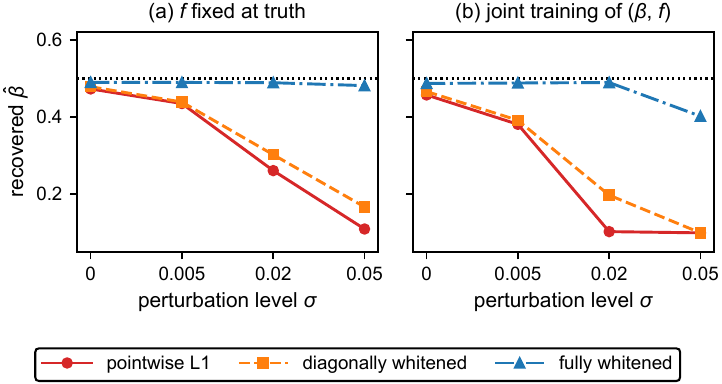}
\caption{Recovered $\hat{\beta}$ versus noise level $\sigma$ (true
$\beta = 0.5$, dotted) for the pointwise, diagonally whitened, and fully whitened objectives. (a) $\beta$-only recovery with $f$ held at its ground truth. (b) Joint recovery of $(\beta, f)$. Under full whitening, $\hat{\beta}$ stays within $0.013$ of the truth up to $\sigma = 0.02$ in both panels, whereas the pointwise and diagonally whitened objectives collapse toward the lower bound.}
\label{fig:whitening}
\end{figure}

Diagonal whitening, however, still fails, because it treats the residuals as independent. The memory sum reuses each noisy increment at every later node, so a single error propagates and compounds across all subsequent residuals. Treating the nodes as independent ignores these correlations entirely, and the optimizer exploits this: $\hat{\beta}$ collapses almost as fast as before (Figure~\ref{fig:whitening}).

We resolve this with \emph{full whitening}, which keeps the entire covariance of the residual vector rather than only its diagonal. Where diagonal whitening normalizes each node independently, Eq.~\eqref{eq:whitened} measures the residual vector as a whole in data units. The metric rescales with the operator, so the gain cancels, and $JJ^{\top}$ carries the inter-node correlations. This requires no noise estimate and introduces no new hyperparameters. The objective is
\begin{equation}
\mathcal{L}_{\mathrm{full}}
\;=\; \frac{1}{m}\,
\Big(
\underbrace{A(\beta)\,c^{\mathrm{obs}}}_{\text{L1 scheme}}
\;-\;
\underbrace{\alpha\, f\!\big(c^{\mathrm{obs}}\big)}_{\text{reaction term}}
\Big)^{\!\top}
\underbrace{\big( J J^{\top} \big)^{-1}}_{\text{noise metric}}
\Big( A(\beta)\,c^{\mathrm{obs}} \;-\; \alpha\, f\!\big(c^{\mathrm{obs}}\big) \Big),
\label{eq:whitened}
\end{equation}
where $c^{\mathrm{obs}} = (c^{\mathrm{obs}}_{0},\dots,c^{\mathrm{obs}}_{m})^{\top}$
collects all $m{+}1$ observations, $A(\beta)\in\mathbb{R}^{m\times(m+1)}$
is the matrix form of the L1 operator (row $n$ reproduces the L1 sum of
Figure~\ref{fig:loss_cheat}), $f$ acts elementwise at the interior nodes
$n = 1,\dots,m$, and the two parenthesized factors are the residual
vector $r$ of Eq.~\eqref{eq:diagwhitenpart2}. Expanding
Eq.~\eqref{eq:jacobian} entry by entry (using $b^{(\beta)}_{0}=1$):
\begin{equation}
J_{nk} \;=\; \frac{\partial r_{n}}{\partial c^{\mathrm{obs}}_{k}}
\;=\;
\begin{cases}
\;g(\beta) \;-\; \alpha\, f'\!\big(c^{\mathrm{obs}}_{n}\big),
& k = n,\\[4pt]
\;g(\beta)\,\big(b^{(\beta)}_{n-k} - b^{(\beta)}_{n-k-1}\big),
& 1 \le k \le n-1,\\[4pt]
\;-\,g(\beta)\, b^{(\beta)}_{n-1},
& k = 0,\\[4pt]
\;0, & k > n,
\end{cases}
\label{eq:jacobian_entries}
\end{equation}
with $g(\beta) = h^{-\beta}/\Gamma(2-\beta)$ the operator gain and
$b^{(\beta)}_{j}$ the L1 weights, both as in Figure~\ref{fig:loss_cheat}.

The fully whitened objective works as intended. As shown in Figure~\ref{fig:whitening}, parameter recovery on perturbed synthetic data now succeeds. At $\sigma = 0.02$ the jointly trained model recovers $\hat{\beta} = 0.4895$ with an $f$ correlation of $0.9986$. PySR recovers the correct law from the learned $f$ (correlation $0.99998$), and the trained objective lands within $1\%$ of the floor set by the true parameters. The joint fit degrades at the highest noise level ($\hat{\beta} = 0.40$ at $\sigma = 0.05$), which we attribute to the flexible $f$ absorbing noise structure rather than to the gain exploit: with $f$ held at its true value (Figure~\ref{fig:whitening}a), $\hat{\beta}$ stays within $0.02$ of the truth at every noise level. Based on the results of both experiments, we adopt this objective for training on real data.

\FloatBarrier

\subsection{Data Processing}
\label{sec:data}

X engagement data were collected with a custom Python scraper deployed on an AWS EC2 instance and integrated with the official X API.  New posts were discovered through 88 keyword queries spanning technology, AI/ML, and startups. Reposts were excluded because they typically start with nonzero engagement at the time of reposting. From November 2025 through January 2026 the scraper discovered 9{,}175 unique posts by 7{,}560 distinct authors, of which 5{,}264 received tracking snapshots; the remainder were discovered too late or missed by the script. For each tracked post, four public engagement metrics (likes, comments, reposts, and quote posts) were recorded at hourly intervals over 48 hours (median 47 snapshots per post). These metrics are summed into a single cumulative engagement total, which is the quantity the model learns.

Although posts were tracked for 48 hours, we model only the first 24 (hourly nodes 0--23). Of the 5{,}158 tracked posts with snapshots on both days, 77\% reach their 48-hour engagement total by hour 23 and 84\% come within 5\% of it; among the complete-grid posts we model, the median post has reached 96\% of its 48-hour total by hour 23. The second day therefore contributes little growth signal while adding plateau nodes that conflict with a strictly positive reaction law. A post enters the trainable pool if its first 24 hours form a complete hourly grid (hour 0 observed, gaps of at most two consecutive hours), its engagement is not flat, and its peak engagement reaches at least 25. Small gaps are filled with linear interpolation.

We require at least 25 engagements to cap quantization noise at the level the whitened objective was shown to tolerate in Section~\ref{sec:perturbed}. Engagement arrives in integer counts, so a max-normalized trajectory can only move in steps of
$1/\mathrm{peak}$. A post peaking at 25 moves in 4\% jumps, while one
peaking at 1{,}000 moves in 0.1\% increments. Treating this rounding as per-node measurement noise of size $\sigma_{\mathrm{eff}} \approx 0.58/\mathrm{peak}$ (where $0.58 \approx 1/\sqrt{3}$ is the standard
deviation of an error spread uniformly over $\pm 1$ count), the filter
guarantees $\sigma_{\mathrm{eff}} \le 0.58/25 \approx 0.023$ for every post. In the worst case, a post therefore has about the same noise as the $\sigma \le 0.02$ bound of Section~\ref{sec:perturbed}; every other post lies comfortably inside this range (e.g., the median post, with $\mathrm{peak} = 96$, has $\sigma_{\mathrm{eff}} \approx 0.006$).

These filters yield 503 complete-grid posts, which we split 80/20 into training and test sets, stratified by peak-engagement decile so that both sides span small to viral posts (403 training and 100 test posts). We additionally remove posts whose cumulative engagement decreases at any hourly step; we suspect these to be instances of X removing bot engagement. This no-decline filter removes 55 training and 14 test posts, leaving the 348 training and 86 test posts used throughout the remainder of the paper. Together these 434 posts span peaks of 25 to 17{,}189 engagements (median 96) and authors with 3 to 36.6 million followers (median 12,123). Table~\ref{tab:datasets} summarizes the splits.

\begin{table}[!htb]
\centering\begin{threeparttable}

\caption{Real-world dataset splits. Train and test are a stratified 80/20 split of the 503 complete-grid posts; the no-decline filter is applied after the split.}
\label{tab:datasets}
\centering\begin{tabular}{|cccc|}
\hline
\textbf{Split} & \textbf{Complete-grid posts} & \textbf{After no-decline filter} & \textbf{Role} \\
\hline
Train (seen)   & 403 & 348 & model fitting \\
Test (unseen)  & 100 & 86  & held-out evaluation \\
\hline
\end{tabular}
\end{threeparttable}
\end{table}

\section{Results}

\subsection{Single-Window Baseline: Training}
\label{sec:baseline}

\begin{figure}[!htb]
\centering
\includegraphics[width=\textwidth]{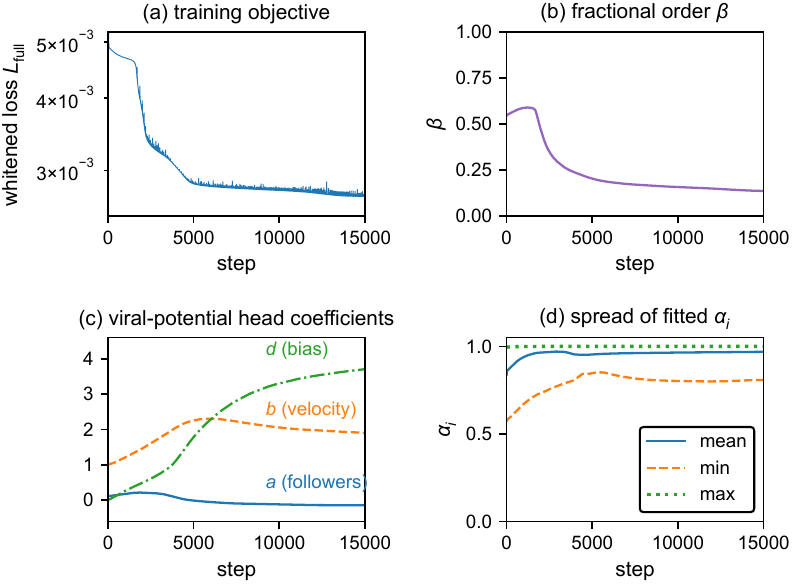}
\caption{Baseline training on real X data. (a) Whitened loss (plotted on logarithmic scale); its square root is the implied per-node observation noise. (b) The fractional order descends to $\hat{\beta} = 0.136$. (c) Coefficients of the viral-potential head, Eq.~\eqref{eq:alpha}: the velocity weight $b$ dominates while the followers weight $a$ decays to ${\approx}\,0$. (d) The fitted potentials $\alpha_i$ remain compressed in $[0.81, 1.00]$.}
\label{fig:baseline_training}
\end{figure}

We now train the model on the real X engagement data. The setup is unchanged from the perturbed synthetic experiment (the same reaction network for $f$, the same $\beta$ parameterization, and the fully whitened objective of Eq.~\eqref{eq:whitened}), with one addition: the per-post viral potential $\alpha_i$, which was held fixed at $1$ throughout the synthetic validation but is now learned from each post through Eq.~\eqref{eq:alpha}. All three components are trained jointly on the 348 training posts for 15{,}000 Adam steps.  Figure~\ref{fig:baseline_training} summarizes the optimization: the loss falls by over an order of magnitude, and every parameter group converges well within the step budget.

These curves reveal three main findings. First, the fractional order converges to $\hat{\beta} = 0.136$ (Figure~\ref{fig:baseline_training}b), implying that engagement is deeply memory-dominated. This is far from the Markovian limit $\beta \to 1$ that classical growth curves assume.  Second, early velocity, not audience size, carries the per-post viral potential. The velocity weight climbs to $b = 1.91$ while the followers weight decays (Figure~\ref{fig:baseline_training}c).  The compressed range $\alpha_i \in [0.81, 1.00]$ additionally implies that per-post differentiation in ``viral potential'' is weak in this single-window model. Third,
the converged loss (Figure~\ref{fig:baseline_training}a) implies a
per-node observation noise of $0.052$. This is not measurement error: the peak filter of Section~\ref{sec:data} caps quantization noise at $0.023$ ($0.006$ for the median post). Since independent errors add in variance, counting noise accounts for under $20\%$ of the residual variance even in the worst case. The remainder is real hour-to-hour variation that a single smooth global law cannot express.

\begin{figure}[!htb]
\centering
\includegraphics[width=9.5cm]{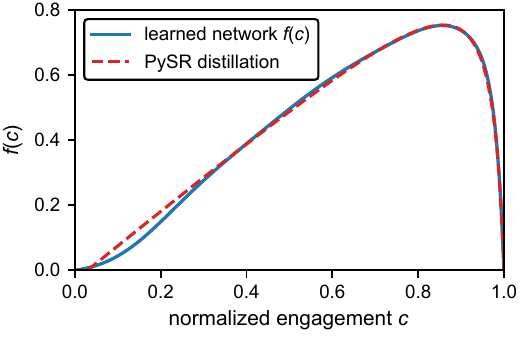}
\caption{Learned reaction law $f(c)$ (solid) and its symbolic distillation, Eq.~\eqref{eq:baseline_law} (dashed), correlation $0.9988$: near-linear growth with a sharp cutoff at saturation rather than a symmetric logistic hump.}
\label{fig:baseline_f}
\end{figure}

We then distill the learned network into closed form with PySR. Figure~\ref{fig:baseline_f} shows the learned reaction law and its symbolic distillation.  The learned $f$ rises almost linearly through the bulk of a post's life, peaks near $c \approx 0.85$, and shuts off sharply at saturation.  The best symbolic candidate is
\begin{equation}
\hat{f}(c) \;=\; 1.097\left(c - \frac{0.0295}{1.0295 - c}\right),
\label{eq:baseline_law}
\end{equation}
which matches the network with correlation $0.9988$. Growth remains proportional to accumulated engagement until close to saturation, then stops abruptly. Together with $\hat{\beta} = 0.136$, the single-window model reads as memory-maximal, near-linear fractional growth.

Assembling the three fitted components gives the complete recovered
governing law for the single-window baseline:
\begin{align}
D_t^{\,0.136}\, c_i(t)
\;&=\; \hat{\alpha}_i
\left[\, 1.097\!\left(c_i(t) \;-\; \frac{0.0295}{1.0295 - c_i(t)}\right) \right],
\qquad c_i(0) = c_{i,0},
\label{eq:baseline_full}\\[2pt]
\hat{\alpha}_i \;&=\;
\sigma\!\big({-0.143}\,\log(1+N_i) \;+\; 1.910\, v_i \;+\; 3.708\big).
\nonumber
\end{align}
\FloatBarrier

\subsection{Single-Window Baseline: Testing}
\label{sec:baseline_test}
\begin{figure}[!htb]
\centering
\includegraphics[width=\textwidth]{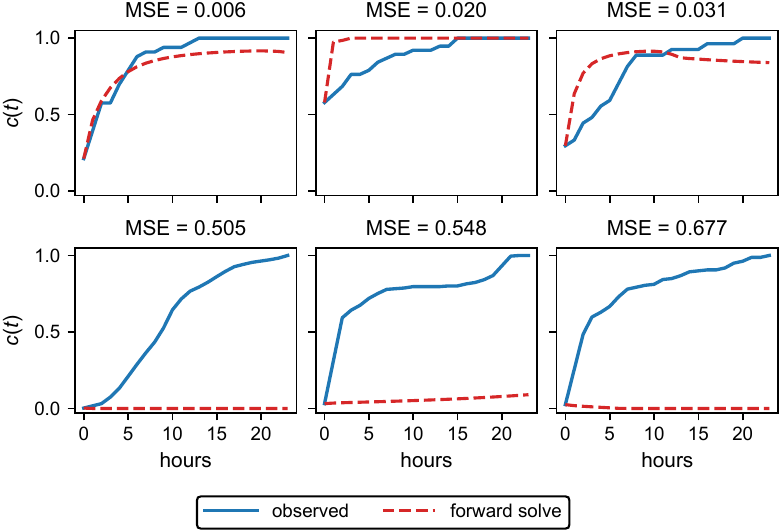}
\caption{Held-out test set forward solves (shape only): observed trajectories (solid) versus the recovered equation solved from each post's $c(0)$ (dashed), for six test posts spanning the 5th--98th percentiles of trajectory MSE. Fast-saturating posts are captured (top row); steady climbers from small initial values are under-forecast (bottom row).}
\label{fig:baseline_forecasts}
\end{figure}

We then test the accuracy of Eq.~\eqref{eq:baseline_full} on the held-out test set. For each post, we forward-solve Eq.~\eqref{eq:baseline_full} from the observed initial value $c(0)$ over the remaining 23 hours and score the trajectory mean squared error (MSE) against the observed curve. Because the curves are per-post max-normalized, this evaluation grades shape only: we ask whether the recovered law reproduces \textit{how} a post climbs to its peak, not \textit{what} the peak is. On the held-out test set, the baseline recovers $64.0\%$, $53.5\%$, and $10.5\%$ of trajectories at MSE thresholds $0.10$, $0.05$, and $0.01$, respectively. These results are underwhelming, particularly as a straight line from $c(0)$ to $1$ achieves $86.0\%$, $64.0\%$, and $14.0\%$ at the same thresholds. Figure~\ref{fig:baseline_forecasts} shows why: fast-saturating posts are reproduced accurately, but a single global $(\beta, f)$ pair forces every post through one front-loaded shape family. Posts that climb steadily from a small initial value are strongly under-forecast, and the model has no way to capture late acceleration.

\FloatBarrier

\subsection{Teacher–Student Forcing Method}
\label{sec:teacherstudent}

Based on the results of our baseline run, we determine the failure mode of Figure~\ref{fig:baseline_forecasts} to be structural. A single fractional order imposes one memory regime on the entire life of a post and implies that ``memory'' remains constant regardless of where a post is in its life cycle. Within the same post, the model is trying to learn a single scalar that simultaneously represents the dynamics at hour 2 (typically high growth) and the dynamics at hour 20 (typically decaying). This is not possible.

\begin{figure}[!htb]
  \centering
  \includegraphics[width=\textwidth]{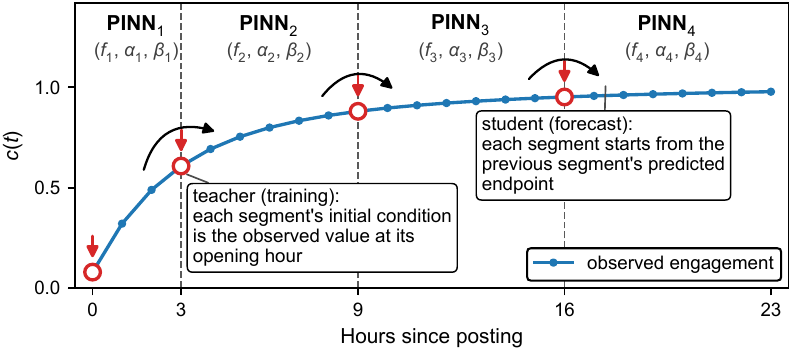}
  \caption{Schematic of the teacher--student forcing method. The window is partitioned into four segments, each fit by an independent PINN with its own $(f_k, \alpha_k, \beta_k)$. During training, every segment's initial condition is anchored to the observed engagement at its opening hour (short vertical arrows: the teacher). At forecast time the segments run as a chain: the first starts from the observed $c(0)$, and each later segment forward-solves from the previous segment's predicted endpoint (curved arrows: the student). The per-segment potentials $\alpha_k$ are computed from the author's follower count and the post's hour-0 to hour-1 velocity.}
  \label{fig:ts_schematic}
  \end{figure}

We therefore update the model architecture and use a teacher--student forcing method (Figure~\ref{fig:ts_schematic}): we partition the window into four segments (hours 0--3, 3--9, 9--16, and 16--23) and train an independent PINN on each segment with its own reaction law $f_k$, viral-potential head $\alpha_k$, and fractional order $\beta_k$. During training, every segment is anchored to the observed engagement at its own opening hour (the teacher); at forecast time the segments are chained, so each takes its initial condition from the previous segment's final predicted endpoint (the student). Beyond the segmented training windows, everything else is the same as in the baseline experiment. The nonuniform segment lengths (3, 6, 7, and 7 hours) are chosen because the first 3 hours of a post tend to have the most variability.

\begin{figure}[!htb]
  \centering
  \includegraphics[width=10cm]{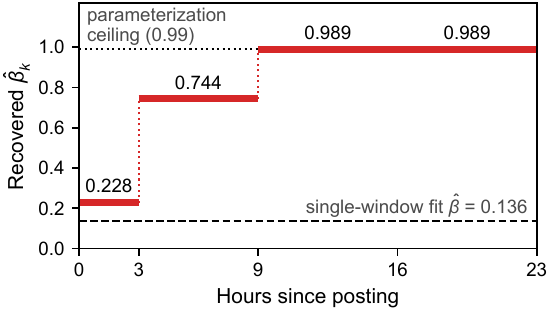}
  \caption{Recovered fractional order per window (values above each segment). The burst window is deeply memory-dominated; the relaxation windows sit at the 0.99 parameterization ceiling (essentially integer-order). The single-window fit (dashed) lies below every segment value: one shared $\beta$ does not average the regimes.}
  \label{fig:ts_betas}
  \end{figure}

The segmented fit is far more effective. As shown in Figure~\ref{fig:ts_betas} and Table~\ref{tab:segments}, $\hat{\beta}$ climbs from $0.228$ in the first window to $0.744$ in the second, and reaches $0.989$ in the final two windows. The implied per-node noise drops from the single-window $0.052$ to $0.023$--$0.043$ per segment, reaching the quantization floor in the final window. The changing $\beta$ values confirm the hypothesis: the single-window model had no way of representing different memory regimes within one post.

\begin{table}[!htb]
\centering\begin{threeparttable}

\caption{Per-segment fits of the teacher--student model. Implied noise is $\sqrt{\mathcal{L}_{\mathrm{full}}}$ for that window (single-window baseline: $0.052$).}
\label{tab:segments}
\centering\begin{tabular}{|cccc|}
\hline
\textbf{Window (h)} & $\boldsymbol{\hat{\beta}_k}$ & \textbf{Implied noise} & \textbf{Distilled law (corr.\ with network)} \\
\hline
0--3   & 0.228 & 0.037 & $1.79\,c + 0.19 + 0.18/(c - 1.09)$ \; (1.000) \\
3--9   & 0.744 & 0.043 & $3.05\,c\,(1 - 0.98\,c)$ \; (0.988) \\
9--16  & 0.989 & 0.031 & $4.77\,c\,(0.98 - c)$ \; (0.668) \\
16--23 & 0.989 & 0.023 & $c\,\big(7.84/(c + 0.29) - 5.81\big)$ \; (0.965) \\
\hline
\end{tabular}
\end{threeparttable}
\end{table}

Symbolic distillation recovers an interpretable law in every window (Table~\ref{tab:segments}).  Read in sequence, the four distilled laws assemble the life cycle of a post: near-linear growth at the beginning, logistic spread in the middle, then saturating decay. Each classical fixed form appears as one \emph{phase} of the dynamics, but no single form occupies all four windows. This is precisely why no single fixed-form curve, classical or fractional, can fit the entire trajectory.

\begin{figure}[!htb]
\centering
\includegraphics[width=\textwidth]{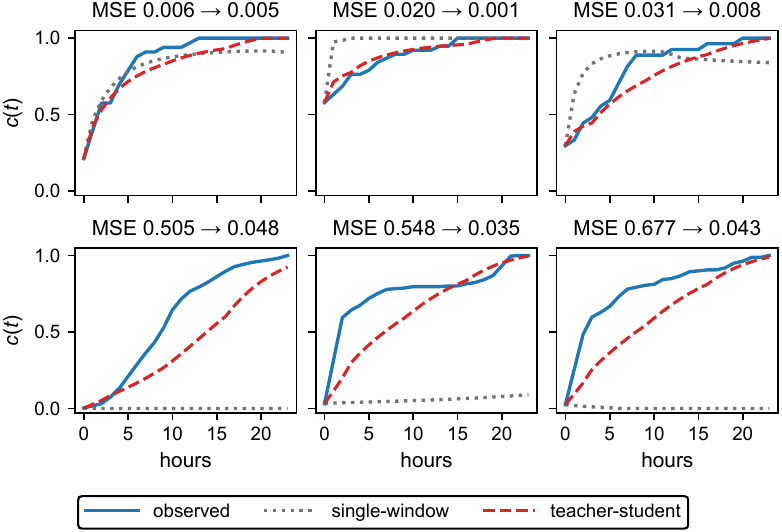}
\caption{Held-out forward solves (shape only) for the same six posts as Figure~\ref{fig:baseline_forecasts}: observed (solid), single-window baseline (dotted), and the teacher--student chain from $c(0)$ (dashed); titles give baseline $\to$ teacher--student MSE.}
\label{fig:ts_forecasts}
\end{figure}

Under the same testing protocol as before (now chaining the four distilled laws from $c(0)$), the teacher--student model recovers $100\%$, $91.9\%$, and $64.0\%$ of the 86 test posts at MSE thresholds $0.10$, $0.05$, and $0.01$, a substantial improvement over the baseline's $64.0\%$, $53.5\%$, and $10.5\%$. Figure~\ref{fig:ts_forecasts} shows the same posts as Figure~\ref{fig:baseline_forecasts}: the steady climbers that the single-window law flatlined on are now tracked closely (MSE $0.505 \to 0.048$, $0.548 \to 0.035$, $0.677 \to 0.043$).

\FloatBarrier

\subsection{Comparison to Non-PINN Baselines}
\label{sec:comparison}

We benchmark the teacher--student PINN against nine classical baselines (eleven counting the follower-scaled logistic and Gompertz variants, which produce identical scores and are folded into their base rows), all fit on the same 348 training posts and all evaluated under the same protocol. The field spans the standard growth curves (logistic, Gompertz, and Bass diffusion), three self-exciting Hawkes processes (exponential, power-law, and learned nonparametric kernels), and three reference predictors (persistence, a linear ramp, and climatology). Table~\ref{tab:baseline_desc} gives a brief description of each baseline.

\begin{table}[!htb]
\centering\begin{threeparttable}

\caption{The classical baselines (the follower-scaled logistic and Gompertz variants are folded into their base rows). All are fit on the 348 no-decline training posts and forward-solved from $c(0)$ under the same protocol as the PINNs.}
\label{tab:baseline_desc}
\centering\begin{tabular}{|cp{8.3cm}|}
\hline
\textbf{Baseline} & \multicolumn{1}{c|}{\textbf{Description}} \\
\hline
Logistic         & symmetric S-curve, growth $\propto c(1-c)$; one global rate $r$ fit to all posts (variant: $r$ scaled by author followers) \\
Gompertz         & asymmetric S-curve whose growth rate decays exponentially; one global decay $b$ (variant: $b$ scaled by author followers) \\
Bass diffusion   & two-parameter adoption model mixing innovation ($p$) and imitation ($q$) \\
Hawkes (exponential)  & self-exciting rate; each engagement's influence fades exponentially \\
Hawkes (power-law)    & self-exciting rate with heavy-tailed, power-law fading influence \\
Hawkes (learned kernel) & self-exciting rate whose kernel shape is learned from the training set by non-negative least squares \\
Persistence      & engagement never moves: $c(t) = c(0)$ for all $t$ \\
Linear ramp      & straight line from $c(0)$ to the peak at hour 23 \\
Climatology      & follows the mean training trajectory from $c(0)$ onward \\
\hline
\end{tabular}
\end{threeparttable}
\end{table}

At the loosest tolerance the benchmark is saturated: every growth curve (and the PINN) recovers at least $95\%$ of held-out posts at MSE $< 0.10$. Table~\ref{tab:comparison} therefore reports the two tighter tolerances. At MSE $< 0.025$ the teacher--student PINN leads, recovering $81.4\%$ of held-out posts against Bass diffusion's $75.6\%$ and Gompertz's $70.9\%$, with the Hawkes family trailing the growth curves throughout.  At the strict MSE $< 0.01$ tolerance the margin widens: $64.0\%$ against $44.2\%$ for Bass, with no other classical model reaching $35\%$. Models are evaluated on both seen (training) and unseen (test) data to check generalization.

\begin{table}[!htb]
\centering\begin{threeparttable}

\caption{Percentage of posts with trajectory MSE below each tolerance, forward-solved from $c(0)$.  Seen $=$ the 348 training posts every model was fit on; unseen $=$ the 86 held-out test posts.}
\label{tab:comparison}
\centering\begin{tabular}{|ccccc|}
\hline
 & \multicolumn{2}{c}{\textbf{Seen (train, $n = 348$)}} & \multicolumn{2}{c|}{\textbf{Unseen (test, $n = 86$)}} \\
\textbf{Model} & $< 0.025$ & $< 0.01$ & $< 0.025$ & $< 0.01$ \\
\hline
\textbf{Teacher--student PINN} & \textbf{87.6\%} & \textbf{72.1\%} & \textbf{81.4\%} & \textbf{64.0\%} \\
Bass diffusion                        & 76.1\% & 49.1\% & 75.6\% & 44.2\% \\
Gompertz                              & 74.1\% & 43.7\% & 70.9\% & 34.9\% \\
Climatology (mean curve)              & 63.5\% & 39.7\% & 66.3\% & 33.7\% \\
Logistic                              & 60.9\% & 27.9\% & 54.7\% & 26.7\% \\
Linear ramp $c(0) \to 1$              & 37.1\% & 16.1\% & 41.9\% & 14.0\% \\
Hawkes (learned kernel)               & 39.4\% & 24.1\% & 36.0\% & 20.9\% \\
Hawkes (exponential)                  & 27.0\% & 15.8\% & 26.7\% & 12.8\% \\
Hawkes (power-law)                    & 20.7\% & 15.2\% & 22.1\% & 14.0\% \\
Persistence ($c = c(0)$)              & 0.9\%  & 0.9\%  & 0.0\%  & 0.0\%  \\
\hline
\end{tabular}
\end{threeparttable}
\end{table}
\FloatBarrier

The table supports the paper's central claim. Classical growth curves are excellent at coarse shape, but each commits to a fixed functional form. Because the teacher--student model also divides the trajectory into four windows, we run two ablations to separate the contribution of segmentation from that of fractional memory. First, we refit each classical growth curve piecewise on the same four windows as in the teacher--student model. Across all baselines, this raises accuracy over their single-window counterparts by at most 8.1\% (Bass diffusion). All classical baselines still trail the teacher--student PINN's $64.0\%$. Second, we retrain the teacher--student model with $\beta_k = 1$ in every window (five seeds per arm), keeping the data, loss, teacher--student forcing, and training budget unchanged. Control runs whose reaction functions remain bounded recover $52$--$62\%$ of held-out posts at MSE $< 0.01$, far above the single-window baseline's $10.5\%$ and Bass's $44.2\%$. To isolate the additional contribution of fractional memory, we compare the learned fractional orders against $\beta=1$ in the first nine hours, holding the later dynamics constant. Removing fractional memory reduces accuracy by about 8\% ($p<0.05$), with the largest effect in the initial 0--3 h window. These results suggest that while segmentation and learned reaction laws account for much of the improvement, fractional dynamics still provide a meaningful benefit.

\section{Conclusions}

This paper introduces a fractional physics-informed framework that discovers the governing law of social media virality on X rather than assuming one.  The model expresses a post's engagement rate --- computed from its full history through a Caputo derivative of learned order $\beta$ --- as the product of a per-post viral potential and a universal reaction law $f$. Synthetic experiments at the 24-node deployment resolution certify that this inverse problem is solvable, even when the synthetic data are perturbed.

Three main findings follow. A single-window fit finds deeply memory-dominated dynamics ($\hat{\beta} = 0.136$) whose per-post signal is early velocity rather than audience size. However, this model is less accurate than traditional growth curves because the dynamics of virality change over a post's life: no single fractional order can capture a post's changing memory dynamics over time. The teacher--student forcing method resolves this by recovering a memory arc ($\hat{\beta}$: $0.228 \to 0.744 \to 0.989 \to 0.989$; burst, transition, relaxation, relaxation) and a distinct interpretable phase law for each window. Combined, this method achieves an accuracy that no baseline matches, reproducing $64.0\%$ of held-out posts within MSE $0.01$, a ${\sim}45\%$ relative improvement over the next best classical model (Bass diffusion). Strong performance on unseen data additionally rules out pure memorization.

\subsection{Limitations}

Our conclusions are bounded by the data. The final sample of 348 training and 86 test posts is small relative to the daily volume of new posts on X, which has roughly 200 million daily active users. Additionally, all posts were discovered through 88 keyword queries concentrated in technology, AI/ML, and startups. Different topics may follow different virality dynamics, so the recovered constants apply only to one small subset of X. Virality on X is additionally governed by a recommendation algorithm that not only differs from those of other platforms (so results cannot be transferred to other social media platforms), but is frequently updated. Every revision to X's algorithm may shift the memory structure and growth law, so our parameters hold only within the November 2025--January 2026 collection window.

The model itself is also deliberately scoped. We predict only shape conditional on scale, not actual magnitude. In other words, we pay no attention to ``what will go viral'' and focus only on ``what trajectory will it take.'' We hope to eventually train a prediction model that can forecast final engagement counts.

\subsection{Future Work}

First, the teacher--student segmentation is a piecewise-constant approximation of what is more naturally a variable-order operator. Modeling $\beta$ as a learned function $\beta(t)$ would allow memory to evolve continuously over a post's life, eliminating the hand-chosen window boundaries and per-segment memory restarts. In this structure, $\beta$ could be represented by a second neural network (similar to $f$) and distilled with PySR. $\beta$ could even be represented by a multivariable function $\beta(t, \epsilon)$, where $t$ is time and $\epsilon$ is some other conditioning variable that affects memory dynamics (e.g., the topic of the post).

Second, this paper deliberately addresses only one half of the ``predicting social media virality'' problem: the shape of growth, never the final scale. Pairing the recovered shape law with a separate model that predicts final engagement would complete the end-to-end forecaster.

Lastly, we hope our research encourages others to apply this framework to other social media platforms, particularly ones with more regimented and observable interaction dynamics. On X, virality is feed-driven and engagement patterns cannot be reconstructed from an explicit graph, which is why we use a fractional derivative term instead of a graph Laplacian term. On Reddit, by contrast, content spreads across an explicit community structure (subreddits). This supplies a natural graph Laplacian, making the traditional reaction--diffusion form more feasible. More broadly, as data in varying areas of social science become more available, we hope to see more novel applications of PINNs.


\acknowledgements

This work was supported by the Office of Naval Research (ONR) through Vannevar Bush Faculty Fellowship Grant No.~N00014-22-1-2795, ``The Next Generation of Operator Regression Networks: Theory, Algorithms, Applications.''

\section*{DECLARATION OF GENERATIVE AI USE}

During the preparation of this work the authors used Claude (Anthropic) in order to improve the readability and language of the manuscript. After using this tool, the authors reviewed and edited the content as needed and take full responsibility for the content of the publication.

\section*{DATA AVAILABILITY STATEMENT}

The dataset analyzed in this paper consists of hourly cumulative engagement counts (likes, comments, reposts, and quote posts) over the first 24 hours after posting for the 503 complete-grid posts on X described in Section~\ref{sec:data}. The processed trajectories, split manifest, and code are publicly available at \url{https://github.com/alexanderkoong/fractional-pinns}.











\bibliographystyle{Bibliography_Style}

\bibliography{References}

@article{raissi2019pinn,
  author  = {M. Raissi and P. Perdikaris and G. E. Karniadakis},
  title   = {Physics-informed neural networks: A deep learning framework for solving forward and inverse problems involving nonlinear partial differential equations},
  journal = {J. Comput. Phys.},
  volume  = {378},
  pages   = {686--707},
  year    = {2019}
}

@article{karniadakis2021physics,
  author  = {G. E. Karniadakis and I. G. Kevrekidis and L. Lu and P. Perdikaris and S. Wang and L. Yang},
  title   = {Physics-informed machine learning},
  journal = {Nat. Rev. Phys.},
  volume  = {3},
  number  = {6},
  pages   = {422--440},
  year    = {2021}
}

@article{pang2019fpinns,
  author  = {G. Pang and L. Lu and G. E. Karniadakis},
  title   = {\mbox{fPINNs}: Fractional physics-informed neural networks},
  journal = {SIAM J. Sci. Comput.},
  volume  = {41},
  number  = {4},
  pages   = {A2603--A2626},
  year    = {2019}
}

@article{zhang2024discovering,
  author  = {Z. Zhang and Z. Zou and E. Kuhl and G. E. Karniadakis},
  title   = {Discovering a reaction\textendash{Diffusion} model for {Alzheimer's} disease by combining {PINNs} with symbolic regression},
  journal = {Comput. Methods Appl. Mech. Eng.},
  volume  = {419},
  pages   = {116647},
  year    = {2024}
}

@article{cranmer2023pysr,
  author  = {M. Cranmer},
  title   = {Interpretable machine learning for science with {PySR} and {SymbolicRegression.jl}},
  journal = {arXiv:2305.01582},
  year    = {2023}
}

@article{caputo1967linear,
  author  = {M. Caputo},
  title   = {Linear models of dissipation whose {Q} is almost frequency independent {II}},
  journal = {Geophys. J. R. Astron. Soc.},
  volume  = {13},
  number  = {5},
  pages   = {529--539},
  year    = {1967}
}

@article{lin2007finite,
  author  = {Y. Lin and C. Xu},
  title   = {Finite difference/spectral approximations for the time-fractional diffusion equation},
  journal = {J. Comput. Phys.},
  volume  = {225},
  number  = {2},
  pages   = {1533--1552},
  year    = {2007}
}

@article{diethelm2002predictor,
  author  = {K. Diethelm and N. J. Ford and A. D. Freed},
  title   = {A predictor-corrector approach for the numerical solution of fractional differential equations},
  journal = {Nonlinear Dynamics},
  volume  = {29},
  number  = {1--4},
  pages   = {3--22},
  year    = {2002}
}

@article{verhulst1838notice,
  author  = {P.-F. Verhulst},
  title   = {Notice sur la loi que la population suit dans son accroissement},
  journal = {Correspondance Math\'{e}matique et Physique},
  volume  = {10},
  pages   = {113--121},
  year    = {1838}
}

@article{gompertz1825nature,
  author  = {B. Gompertz},
  title   = {On the nature of the function expressive of the law of human mortality, and on a new mode of determining the value of life contingencies},
  journal = {Philos. Trans. R. Soc. Lond.},
  volume  = {115},
  pages   = {513--583},
  year    = {1825}
}

@article{bass1969diffusion,
  author  = {F. M. Bass},
  title   = {A new product growth model for consumer durables},
  journal = {Manage. Sci.},
  volume  = {15},
  number  = {5},
  pages   = {215--227},
  year    = {1969}
}

@article{hawkes1971spectral,
  author  = {A. G. Hawkes},
  title   = {Spectra of some self-exciting and mutually exciting point processes},
  journal = {Biometrika},
  volume  = {58},
  number  = {1},
  pages   = {83--90},
  year    = {1971}
}

@inproceedings{zhao2015seismic,
  author    = {Q. Zhao and M. A. Erdogdu and H. Y. He and A. Rajaraman and J. Leskovec},
  title     = {{SEISMIC}: A self-exciting point process model for predicting tweet popularity},
  booktitle = {Proc. 21st ACM SIGKDD Int. Conf. Knowl. Discovery Data Mining},
  pages     = {1513--1522},
  year      = {2015}
}

@book{fuller1987measurement,
  author    = {W. A. Fuller},
  title     = {Measurement Error Models},
  publisher = {John Wiley \& Sons},
  address   = {New York},
  year      = {1987}
}

@inproceedings{kingma2015adam,
    author    = {D. P. Kingma and J. Ba},
    title     = {Adam: A method for stochastic optimization},
    booktitle = {Proc. 3rd Int. Conf. Learn. Represent. (ICLR)},
    address   = {San Diego, CA},
    year      = {2015}    }
\end{document}